\documentclass{article}
\usepackage{graphicx}

\usepackage{arxiv_upload}

\usepackage[utf8]{inputenc} 
\usepackage[T1]{fontenc}

\usepackage{soul}

\usepackage{hyperref}       
\usepackage{url}            
\usepackage{booktabs}       
\usepackage{amsfonts}       
\usepackage{nicefrac}       
\usepackage{microtype}      
\usepackage{xcolor}         

\providecommand{\ywreply}[2]

\usepackage{times}
\usepackage{nicefrac}
\usepackage{mathtools}
\usepackage{amsthm}
\usepackage{amsmath}
\usepackage{amssymb}
\newcommand{\ie}{\emph{i.e.}}
\newcommand{\eg}{\emph{e.g.}}
\newcommand{\wrt}{\emph{w.r.t.}}

\newcommand{\expectation}{\mathbb{E}}

\newtheorem{assumption}{Assumption}
\newtheorem{lemma}{Lemma}

\newtheorem{theorem}{Theorem}

\newtheorem{definition}{Definition}
\usepackage{algorithm}
\usepackage{algpseudocode}

\title{Compositional Hardness of Code in Large Language Models - A Probabilistic Perspective}
\author{Yotam Wolf, Binyamin Rothberg$^*$,~~~~ \kern-1em Dorin Shteyman\thanks{Equal contribution}~~, and Amnon Shashua
\\The Hebrew University
\\\texttt{\{yotamwolf,binyamin.rothberg,dorin.shteyman,shashua\}}
\\\texttt{@cs.huji.ac.il}}

\date{September 2024}

\begin{document}

\maketitle
\begin{abstract}
A common practice in large language model (LLM) usage for complex analytical tasks such as code generation, is to sample a solution for the entire task within the model's context window. Previous works have shown that subtask decomposition within the model's context (chain of thought), is beneficial for solving such tasks. In this work, we point a limitation of LLMs' ability to perform several sub-tasks within the same context window -- an in-context hardness of composition, pointing to an advantage for distributing a decomposed problem in a multi-agent system of LLMs. The hardness of composition is quantified by a generation complexity metric, \ie, the number of LLM generations required to sample at least one correct solution. We find a gap between the generation complexity of solving a compositional problem within the same context relative to distributing it among multiple agents, that increases exponentially with the solution's length. We prove our results theoretically and demonstrate them empirically.
\end{abstract}

\section{Introduction}
Large language models (LLMs), based on the transformer archietecture \citep{vaswani2017attention}, have become very efficient problem solvers in many domains, such as broad-scoped question answering, 
writing assistance, teaching, and more \citep{brown2020language,radford2019language,openai2023gpt,bubeck2023sparks,nori2023capabilities,west2023advances}.
Yet their analytical skills, such as coding capabilities, were slow to develop - \cite{chen2021evaluating,li2022competition,AlphaCode2T,ridnik2024code} showed LLMs struggle on competitive coding problems, and ~\cite{zhuo2024bigcodebench} showed they struggle on complex coding tasks. 

One way to empower LLMs in analytical tasks, is to use subtask decomposition, otherwise known as chain of thought (CoT) -- a method in which an LLM breaks down a problem to smaller, more manageable tasks, solves them, and integrates it into a solution. The method has been empirically demonstrated by \cite{wei2022chain}, that show reasoning capabilities of language models improve when they are prompted to break down a task. Since then, the method has been incorporated into the training phase of LLMs, in models such as openAI's o1 \cite{jaech2024openai}, Deep-Seek-R1 \cite{guo2025deepseek}, QWQ \cite{teamqwq} and Kimi K1.5 \cite{team2025kimi}, achieving state of the art results on existing coding benchmarks. Its efficiency has also been studied theoretically -  \cite{wies2022sub,malach2023auto,merrill2023expresssive,sanford2024representational} proving through the autoregressive nature of language models, that problems that cannot be solved directly, can be solved by subtask decomposition, allowing to represent any polynomial time turing machine with a polynomial number of CoT steps.

Yet even with task decomposition, due to the transformer architecture's limited ability to compose functions and learn complex relations between tokens  \cite{sanford2024representational,peng2024limitations,xu2024large}, some tasks require an arbitrarily long CoT for an LLM to solve. This is especially prominant in SOTA LLMs such as openAI's o3, that generates $\sim 100k$ tokens per problem on the ARC AGI benchmark \cite{chollet2024openai}.
However, in practice, LLMs are limited in their context length -- beyond the constraint of context length during training, \cite{hsieh2024ruler,liu2024lost} show that in practice many models can perform tasks only on a much shorter context length than they were trained on and cannot fully use all the information within their context. Consequently, even though CoT can in theory allow an LLM to solve arbitrarily complex analytical problems, in practice, they will be limited by the effective context length. This is especially concerning for more practical programming applications than those considered in coding benchmarks, such as building and interacting with large code bases, too big for any LLM context to contain.

A rising approach to remedy this limitation is to solve problems through the use of multi-agent systems, that tackle complex problems through the use of agents, where each agent is an LLM instance that solves a different aspect of the problem. While it has been used for simulating social interactions, \citep{park2023generative,li2023metaagents,pang2024self}, it has also been shown as an effective tool for analytical problem solving. This can be done by decomposing a large task and distributing the sub-tasks between agents. \cite{ishibashi2024self} use this method for building large code bases and \cite{liu2023dynamic}, use an LLM-agent network for solving code problems and analytical tasks.

In this work, we theoretically study a compositional hardness of coding problems originating from context processing limitations of LLMs, and the resultant effectiveness of a multi-agent system over a single model instance in composite coding problems. We model a coding problem as a composition of simpler sub-problems, as is typically invoked by LLMs in a CoT process, such that combining the codes to all the sub-problems creates a program that implements the full solution. The model's usefulness on a coding task is quantified by a generation complexity metric (definition \ref{def:gen_comp}) -- the number of LLM generations required to sample at least one correct solution. The appeal of this metric is that due to the existence of code testing units, it suffices to turn an LLM into a good program candidate generator and simply output a candidate that is correct \citep{kambhampatiposition,thakur2023language,luo2024improve}. When a single LLM instance solves all the sub-problems in a single context, its usefulness is the generation complexity for the composite problem. The multi-agent on the other hand, distributes the sub-problems between LLM instances, sampling a corresponding solution from each, its helpfulness is the product of generation complexities of the sub-problems, corresponding to sampling independently a correct solution to each.

We theoretically model an LLM as an autoregressive model, where a solution is sampled token by token, based on the hidden representations. When combining two problems whose solutions are grammatically similar but semantically different (such as different coding problems) into a single context window, we assume this combination injects noise into the model's representations during solution generation for each sub-problem (assumption \ref{assumption:1}), which we denote as screening. We show that a compositional problem may have an exponential gap in generation complexity relative to the product of sub problems' generation complexities (theorem \ref{theorem:1}), meaning an exponential hardness of composition in-context. Essentially, the model is less capable of solving two problems if they are presented within the same context, than if presented in separate contexts. This points to an advantage of decomposing a problem not only within the same LLM context, but to distribute the problems among multiple agents (\ie~solve each sub-task within a different context).
Additionally, this result provides a view of the model's effective context length through the lens of screening, which is the model's ability to isolate the relevant context at each decoding step -- additional irrelevant context may reduce model's performance on other tasks within the context window exponentially with length.
We validate our assumptions and results experimentally on Llama 3 8B Instruct and 70B Instruct, and Mistral 7B Instruct V0.3 by constructing composite code problems.

\section{Related Works}

\paragraph{LLMs as solution candidate generators in programming:} Past empirical works have shown that while LLMs struggle to solve complex programming tasks with a single sampled solution, on some problems sampling thousands of solutions may yield a handful of correct ones that can be detected by testing units \citep{chen2021evaluating,li2022competition,AlphaCode2T,ridnik2024code}. Hence empowering an LLM with a large sampling budget drastically boosts its capabilities. This procedure has of recent gained popularity under the term "scaling test time compute", which aims at leveraging compute at inference time \cite{snell2024scaling}, showing it may be more advantageous than using it for additional training time. This has been recently demonstrated at the extreme scale with the results of OpenAI's o3 on the ARC AGI challenge \cite{chollet2024openai}, where an increased budget of 1024 parallel samples per problem from the model contributed to a significantly higher performance than using 6 parallel samples. As such, a relevant approach to quantify an LLM's usefulness in such tasks is the expected number of programs one needs to sample from the model to obtain a correct solution. This motivates our definition of generation complexity (definition \ref{def:gen_comp}), and in this work, we primarily focus on providing theoretical results on the relative generation complexity in compositional coding tasks.

\paragraph{Theoretical results on composition:} Previous works on multi-step function composition and subtask decomposition in language models primarily focus on expressing and learning a deterministic function that solves well defined mathematical tasks \cite{ahuja2024provable}, such as solving the bit parity problem \citep{wies2022sub}, expressing and learning Turing machines \citep{merrill2023expresssive,malach2023auto}, composing a mathematical function arbitrary numbers of times \citep{peng2024limitations}, or simple mathematical operations on text \cite{xu2024large}. These do not translate in a natural way to contemporary code generation methods using LLMs, which are based on probabilistic sampling in order to explore the vast solution space, as in the above works, the model provides a deterministic answer which is either correct or incorrect and the notion of sampling multiple solutions cannot be incorporated. Hence beyond the fact that SOTA LLMs do not deterministically generate solutions to coding problems, the importance of scaling test time compute is absent. In this work, we deal with probabilistic autoregressive generation of solutions to problems, which is more applicable to the practice of LLM code generation. We provide a softer and more informative result for \textit{how hard} it is to compose problems with the generation complexity metric, which is not possible through previous approaches and incorporate the existence of multiple correct solutions to a single problem, which is typical in programming.

\paragraph{Effectiveness of LLMs in utilizng long context:} Previous empirical works \citep{hsieh2024ruler,liu2024lost} have shown LLM performance on tasks such as retrieval degrade when performed on longer contexts, and that models may be limited in their random access to tokens within the context \citep{ebrahimi2024your}. In this work, we study the degradation of language models on more complex tasks that involve not just retrieval operations from a given long context, but also include long text generation: Namely, compositional coding tasks through the lens of noise that pieces of context from different subtasks insert into the generation process. Differently from above mentioned works, in our results the context does not necessarily have to be very long in order for the model performance to drop, but rather that grammatically similar but semantically different pieces of the context may ``confuse" the model and harm code generation even in shorter contexts.

\section{Framework}
\subsection{Generation Complexity}\label{sec:gen_comp}
We focus on coding problems, meaning each problem is written in natural language and is solved by a function, and the goal is for the model to generate a code implementing it.
\begin{definition}\label{def:correct_program}
    Let $L$ be a programming language. Let $x$ be a natural language description of a problem, that is solved by function $f$, then a computer program $y$ is a correct solution to $x$, if it implements $f$.
\end{definition}
We formally define the generation complexity of a problem as the inverse success rate of a model to generate a correct solution to the problem:
\begin{definition}\label{def:gen_comp}
    For a problem $x$ and a natural language distribution $P$ over $V^*$, the generation complexity of $x$ \wrt~$P$, is:
    \begin{equation}
        N(P,x) = \frac{1}{\sum_{y\in \textit{correct solutions}}P(y|x)}
    \end{equation}
    Where $V^*$ is the Kleene closure of the vocabulary $V$.
\end{definition}
Intuitively, the generation complexity is the number of program candidates needed to be sampled from the conditional distribution $P(\cdot|x)$ to get a program that solves the problem.

\paragraph{Emergence of compositional coding problems from general coding problems in LLMs with CoT: }In this work, we are interested in deriving generation complexity bounds for coding problems. Generally, LLMs solve coding problems by producing a CoT that suggests a decomposition of the problem $x$, to $k$ subproblems $x_1...x_k$. Each subtask is then implemented by the model as a block of code $y_1...y_k$, and the concatenation of these blocks $y_1\oplus...\oplus y_k$ creates a single program that solves the problem. We note that the different blocks can interact, hence generally it is a compositional problem with $k$ components. Thus all but the simplest coding problems are mapped by LLMs into compositional coding problems of the form described above.

\paragraph{Decoupling hardness of decomposition from hardness of composition in general problems:}
Based on the above, the success rate of a model to generate a correct solution to a problem is determined by two factors. The first is decomposition -- generating a CoT $x_1…x_k$ with a correct decomposition to the problem. The second is composition -- to implement a program $y_1\oplus…\oplus y_k$ matching the subtasks described in the CoT. 
 In the following, we isolate the contribution of composition to generation complexity and show it bounds the generation complexity to the full problem.

Denote the set of valid decompositions to the problem $x$ as:
$d(x)=\{x_1,…,x_k |x_1,…,x_k  \textit{ is a valid decomposition to x} \}$
and the set of correct solutions to a subtask $x_i$ as:
$s(x_i )=\{y|\textit{y is a correct solution to $x_i$}\}$. The probability that a model generates a correct solution to the problem equals the probability to generate a correct decomposition $P(x_1...x_k|x)$, times the probability to implement a correct solution based on the decomposition, which in terms of generation complexity is $\frac{1}{N(P,(x,x_1...x_k))}$. We arrive at the following formula for generation complexity to a general problem as a weighted sum of compositional problems:
\begin{equation}
    \frac{1}{N(P,x)}=\sum_{(x_1...x_k)\in d(x)}\frac{P(x_1,...,x_k|x)}{N(P,(x,x_1...x_k))}
\end{equation}
We see the manifestation of hardness of searching for a decomposition in the numerator, and of implementing a correct solution to a compositional problem in the denominator.
The compositional aspect is interesting since models are famously not good at ignoring irrelevant information, which can harm the ability to solve multiple sub-tasks within the same context. Hence even given a correct decomposition to the problem, implementing it could still prove a challenge. We thus lower bound the generation complexity to the problem in terms of the compositional hardness:
\begin{equation}
    N(P,x)\geq \min_{(x_1...x_k)\in d(x):P(x_1...x_k|x)>0} N(P,(x,x_1...x_k))
\end{equation}
Meaning the generation complexity is at least as large as the generation complexity of solving the decomposed problem for the best decomposition that the model can suggest.
In our main results, we find a non-trivial upper bound on the generation complexity for compositional problems:
\begin{equation}
    N(P,x_1...x_k) \geq [\prod_{i=1}^k N(P,x_i)]\cdot e^{\Delta\cdot \sum_{i=1}^k L_i}
\end{equation}

For some $\Delta>0$, where $L_i$ is the minimal length solution to $x_i$. In words, this result implies that when sampling a solution to sub-problems in one context, $y_1...y_k\sim P(\cdot|x_1...x_k)$, the probability for them all to be correct is exponentially lower (in total code length) than if we sample them in parallel contexts: $y_1\sim P(\cdot|x_1 )...y_k\sim P(\cdot|x_k)$. The implication of this result is that if we do not have access to tests for the intermediate steps $y_1...y_k$, but only for the final output $y_1\oplus...\oplus y_k$ (which is often the case as the model chooses how to decompose a problem), it is more beneficial to sample $y_1...y_k$ in parallel. This defines a compositional hardness of coding problems that originates from context processing limitations.

We note this result is not equivalent to showing a model's success rate in compositional problems drops exponentially with the number of compositions, which was shown in \cite{dziri2024faith}, but rather it is a stronger result. Exponentially decreasing success rate in number of compositions means $N(P,x_1...x_k)\geq \exp(c\cdot k)$ for some $c>0$. But the product of success rates of performing the subtasks in parallel $\prod_{i=1}^k N(P,x_i)$ is naturally also exponential in $k$. Hence such results do not imply anything about whether it is more beneficial to sample in one context or to parallelize. In our results, we show their ratio is exponentially growing, $N(P,x_1...x_k)/(\prod_{i=1}^k N(P,x_i))\geq e^{\Delta\cdot\sum_{i=1}^k L_i}$, which does reveal the advantage to sample in parallel.

\subsection{Screening in Autoregressive Models}
In the above, we have motivated to look at LLM's abilities to solve multiple tasks in a single context compared with solving them in separate contexts. Here we lay theoretical grounds to explain the difference between the two scenarios based on the autoregressive nature of LLMs. For ease of presentation, below we present our assumptions for compositions of two problems, but note that it can inductively be increased to any number. 

Typically, latent representations of the model contain information about the context beyond next token prediction, \eg~structure of the solution to problems \citep{ye2024physics}. Thus when composing two code problems, we expect that during generation, the representations of the second program to contain information about the first and vice versa, which is grammatically similar (same programming language) but semantically very different. As a result, this additional information creates noise that can harm the generation process.

Formally, we denote by $r^{(L)}(x)$ the model's last hidden layer representation of the prompt $x$, by $U$, the model's unembedding matrix (hidden dimension to vocabulary). The logit of the $i$'th token is thus defined as the token's score before the softmax operation: $\langle r^{(L)}(x),U^Te_i\rangle$, where $e_i$ is the one-hot vector of the token. The probability distribution at each decoding step is the softmax applied to the logits, $P_{LLM}(i|x)=softmax(\langle r^{(L)}(x),U^Te_i\rangle)$.

In the process of generating a solution to a compositional code problem, $x$, that is implicitly or explicitly decomposable to $x_1$ and $x_2$, the model will implement a solution to the first part $y_1$ and then to the second part $y_2$. The sequence is generated based on the hidden representations. Informally, we expect the representation of the solution to the first problem, $y_1$, within the compositional problem, $x$, to be a noisy version of the solution's representation in the non-compositional problem, $x_1$:
\begin{equation}
    r^{(L)}(x\oplus y_1)=r^{(L)}(x_1\oplus y_1) + noise
\end{equation}
Similarly, the representation of the second problem's solution, $y_2$, in the compositional problem, $x$, is expected to be a noisy version of the solution's representation in the non-compositional problem, $x_2$:
\begin{equation}
    r^{(L)}(x\oplus y_1\oplus y_2)=r^{(L)}(x_2\oplus y_2) + noise
\end{equation}

Essentially, this means the model attempts to generate the same solutions as in the non-compositional case, but noise may interfere in the process. The projection of this noise onto the dictionary creates noise in the logits during decoding, which can lead the model to make mistakes. It is worth mentioning that while theoretically after generating $y_1$, it may serve as an in-context example for $y_2$, contradicting the noise argument, in practice when composing two complex problems that are semantically different, the model cannot learn from the first problem how to improve its success rate on the second. This remains true for problems decomposed with a chain of thoughts, in which each sub-task serves a different purpose and implementing it requires different tools, hence success in previous sub-tasks are not expected to create an in-context boot-strapping effect.

As the two problems $x_1,x_2$ and their solutions $y_1,y_2$ may be different semantically, we do not expect the noise to ``push" the model towards correct solutions more than to incorrect solutions, so when projected onto the vocabulary, $V$, the noise on the logit of a correct next token, $i_{CNT}$, minus the noise on an incorrect token $i$, denoted, $\langle noise , U^T e_{i_{CNT}}\rangle - \langle noise , U^T e_{i}\rangle$, should be symmetric on average. Additionally it should be bounded within some range $[-M,+M]$, as it changes the hidden representation to a finite extent. In practice, we only expect this to be true for the high probability tokens, as the vocabulary $V$ is very large, and some low probability tokens may be systematically enhanced. To avoid this issue, we make our assumptions only on the weighted average of the noise, where the weights are given by the probability mass that the model assigns them. This way, low probability tokens with asymmetric noise or large norms receive low weight and are averaged with the noise of other tokens. Denote by $P(i|c)$ the probability assigned to the $i$'th token given the context $c$. We make our assumptions on the \textit{weighted average of the noise on incorrect token logits minus correct token logits}:
\begin{equation}\label{eq:weighted_noise}
    X=\frac{\sum_{i\in V\backslash\{i_{CNT}\}}P(i|c)\langle noise, U^Te_i - U^Te_{i_{CNT}}\rangle}{\sum_{i\in V\backslash\{i_{CNT}\}}P(i|c)}
\end{equation}

\begin{assumption}\label{assumption:1}
    Denote by $X$ the weighted noise on the logits as defined in equation \ref{eq:weighted_noise}. We assume that at every given decoding step it is a continuous, symmetric random variable and bounded within $[-M,+M]$ for some $M>0$.
\end{assumption}

In experiment subsection \ref{exp:assumptions} we show the noise satisfies these assumptions, with $M\approx 3-4$.

\subsection{Effect of noise on decoding}
While the noise onto the logits, $X$, averages to zero, its effect on the decoding process does not. The probability of each token in a decoding step given context $c$ is changed to:

\begin{equation}
P(i|c)\rightarrow P'(i|c)\leq\frac{P(i|c)}{P(i|c)+(1-P(i|c))e^X}
\end{equation}
The denominator, $P(i|c)+(1-P(i|c))e^X$ can be thought of as a renormalizing term, which redistributes the probability of the tokens. For $X=0$, the token's probability does not change, for $X<0$ it increases and for $X>0$ it decreases. See appendix \ref{probability_bound} for derivation and intuition. Note that if $P(i|c)\in (\epsilon,1-\epsilon)$ for $\epsilon>0$ (the model has finite confidence), \textit{on average, the noise decreases the probability of a correct continuation} $P(i|c)$, by a factor $\exp(-\Delta(\epsilon,X))$, where $\Delta$ is the renormalizing term's mean:
\begin{equation}\label{eq:delta}
    \Delta(\epsilon,X):=\expectation_X[\log(\epsilon+(1-\epsilon)e^X)]
\end{equation}
Intuitively $\Delta$ is the average renormalization of the correct token's probability. In experiment subsection \ref{exp:assumptions}, we calculate $\Delta(\epsilon,X)$ empirically as a function of $\epsilon$, and find that for $\epsilon=0.1$ for example, $\Delta\approx 0.2$.
The consequence of this, is that on average, most long sequences have their probability reduced by the noise, while few random long sequences have their probability greatly enlarged. Since for long coding problems most sequences are incorrect, the probability of a correct solution getting enlarged is small. We formally show this in the next section. To do so, we will use concentration inequalities, for which we note that:
\begin{equation}\label{eq:M}
    |\log(\epsilon+(1-\epsilon)e^X)|<M
\end{equation}
Thus the renormalizing term's variance is also bounded:
\begin{equation}\label{eq:sigma}
    \sigma^2(\epsilon,X):=Var_X[\log(\epsilon+(1-\epsilon)e^X)]\leq  M^2
\end{equation}

\section{Results}
Here we show that composing coding problems can be significantly harder than solving them in parallel contexts. This helps us derive generation complexity bounds for general problems in terms of simpler problems the model knows how to solve. We present results for composition of two problems, but note it can inductively be increased to any number. A natural quantification of compositional hardness is the ratio between generation complexity of the problem's components and the complete problem. 
\begin{equation}
    \frac{N(P,(x_1,x_2))}{N(P,x_1)\cdot N(P,x_2)}
\end{equation}
A composition is hard if this ratio is significantly larger than $1$, while easy if it is close to $1$.
The rational, is that $N(P,x_1)\cdot N(P,x_2)$ is the number of attempts required to sample in parallel a correct solution to $x_1$ and $x_2$, while $N(P,(x_1,x_2))$ is the number of attempts required to sample a solution to both in a single context. When composition is easy, the model solves the each sub-problem to the best of its abilities, but when composition is hard, seeing both problems combined reduces its performance on each sub-problem, and a multi-agent approach is more favorable.

As there are typically more incorrect solutions to coding problems than correct ones, the random noise inserted into the logits generally harms the model's performance. The following lemma quantifies an exponential decrease in the model's probability of a correct solution to a compositional problem, $P(y_1\oplus y_2|x)$, relative to the probabilties of the sub-problem solutions $P(y_1|x_1)\cdot P(y_2|x_2)$:

\begin{lemma}\label{lemma:1}
Let $\epsilon,\delta\in(0,1)$, and $M> 0$. Let $x$ be a compositional problem and $y_1 \oplus y_2$ a solution, with $x_1$, $x_2$ being the corresponding sub-problems. Suppose that the noise injected to the logits as defined in equation \ref{eq:weighted_noise}, satisfies assumption \ref{assumption:1}, and that the probability assigned to the correct token at each decoding step is bounded within $[\epsilon,1-\epsilon]$. Then there exist strictly positive noise dependent constants $\Delta$ (as defined in equation \ref{eq:delta}) and $c(\Delta,M,\sigma)$ (with $M$ and $\sigma$ as defined in equations \ref{eq:M} and \ref{eq:sigma}), such that if the solution length satisfies $|y_1|+|y_2|>c\ln\frac{1}{\delta}$ 
we have with probability of at least $1-\delta$ (over noise randomness) that:
\begin{equation}
    P(y_1\oplus y_2|x)\leq P(y_1|x_1)\cdot P(y_2|x_2)e^{-\frac{\Delta \cdot(|y_1|+|y_2|)}{4}}
\end{equation}
Where $P(y_1\oplus y_2|x)$ is the probability of producing the answer $y_1\oplus y_2$, given context $x$. The constant $c=\frac{M^2}{\sigma^2\cdot h(\frac{3\Delta\cdot M}{4\sigma^2})}
$, with $h(x)=(x+1)\log(x+1)-x$.
\end{lemma}
The proof is presented in appendix \ref{proof_lemma}. The intuition behind this result is that as there are typically more incorrect choices to make when generating code, random noise usually reduces the probability for sequences with finite confidence. Thus most sequences get their probability reduced, while very few random sequences get a large increase, and these are usually not correct solutions.

The assumption on bounded probability for the correct token $[\epsilon,1-\epsilon]$ implies we are considering solutions where the model has high but limited confidence in each decoding step. While LLMs are often very confident in ``obvious" next steps (line break, etc.), in practice, during generation, nucleus sampling is commonly used, where sampling only occurs if the model is not overly confident. \eg~with $p=0.95$, if a token's probability $P$, is larger than $0.95$, then the probability is rounded $1$. Thus when considering probabilities of sequences, it suffices to look only at decoding steps where the model is not too confident, hence we can consider $\epsilon\approx 0.05$.

In subsection \ref{exp:assumptions}, we see empirically that for $\epsilon=0.1$, $\Delta\approx 0.2, \sigma\approx 1.5, M\approx 4$, making $c\approx 200$, thus for solutions with length $>200$ tokens, the results apply.

In practice, there may be multiple solutions to the same problem (\eg~different implementations). So it is necessary to take all of them into account when considering the generation complexity. 
With this taken into account using a union bound, we obtain the following result of an exponential gap in generation complexity between a composition of problems and the sub-problems, indicating a compositional hardness that is exponential in the solution's length:

\begin{theorem}\label{theorem:1}
    Let $\epsilon,\delta\in (0,1)$, and $N,M>0$. Let $x$ be a compositional problem, with $x_1$, $x_2$ being the corresponding sub-problems. Denote by $L_1, L_2$ the minimal solution length to $x_1, x_2$ respectively, and the total number of solutions to $x$ by $N$. Under the assumptions of lemma \ref{lemma:1}, there exist strictly positive noise dependent constants $\Delta$ (as defined in equation \ref{eq:delta}) and $c(\Delta,M,\sigma)$ (with $M$ and $\sigma$ as defined in equations \ref{eq:M} and \ref{eq:sigma}), such that if the minimal solution length $L_1+L_2$, satisfies $L_1+L_2>c\ln\frac{N}{\delta}$, then with probability of at least $1-\delta$ over the noise's randomness, the generation complexity (definition \ref{def:gen_comp}) satisfies:
    \begin{equation}
        N(P,x)\geq N(P,x_1)N(P,x_2)\cdot e^{\frac{\Delta \cdot (L_1+L_2)}{4}}
    \end{equation}
\end{theorem}
The proof is presented in appendix \ref{proof_theorem}. We see that longer problems become harder to solve within the same context relative to solving them in parallel due to the noise injected into the decoding steps by previously generated tokens. We reiterate this is a stronger statement than the success rate dropping exponentially with the number of compositions, as shown in \cite{dziri2024faith}, but rather that success rate drops exponentially relative to solving the problems in parallel (which in itself is also exponential in number of compositions), leading to a multi-agent efficiency.
This result implies that for long coding problems, it is more beneficial to distribute sub-tasks between different instances of the LLM, and not expose it to the full context.

We note that for the union bound of theorem \ref{theorem:1} to hold, the number of solutions needs to be bounded by an exponential in the length of the solution, $L_1+L_2$:
\begin{equation}
    N<\delta\cdot \exp\left((L_1+L_2)\cdot\frac{\sigma^2}{M^2} h\left(\frac{3\Delta\cdot M}{4\sigma^2}\right)\right)
\end{equation}
Typically, we expect the number of solutions to a problem to grow exponentially relative to shortest solution length, $y$:
\begin{equation}
    N \sim \exp(c\cdot|y|)
\end{equation}
As variable names may be changed, lines implemented differently, etc. Still, the exponential's coefficient is small as most sequences are not correct solutions due to constraints between the tokens (\eg, once a variable name is chosen, it is fixed throughout the solution).
With the empirical values calculated in subsection \ref{exp:assumptions}, we see empirically that the coefficient of the exponential number of solutions is $\approx 0.005$, thus for a solution of length $L_1+L_2=1,600$ tokens, we have $N<\delta \cdot \exp(8) = 3,000\cdot\delta$ solutions.

\subsection{Generation Complexity For General Problems}
Based on the above results for compositional problems, we derive the following lower bound on general code problems for which multiple implicit decompositions may exist:
\begin{theorem}\label{theorem_2}
    Let $x$ be a coding problem. Suppose the model generates a solution by 1) Sampling a decomposition candidate $x_1...x_k$ using CoT, 2) Sampling a solution to the compositional problem $y_1\oplus...\oplus y_k$.  Denote by $\Delta$ and $c$ the constants from theorem \ref{theorem:1} and assume its conditions hold. Denote $d(x)$ the set of all valid decompositions to the problem for which $P(x_1...x_k|x)>0$ (\ie~those the model can generate). If for all $x_1...x_k\in d(x)$ the minimal solution length $L_1...L_k$ satisfies $\sum_{i=1}^k L_i>c \ln(\frac{N\cdot|d(x)|}{\delta})$, where $N$ is the number of solutions to the decomposition, then with probability $1-\delta$, the generation complexity for $x$ is lower bounded by:
    \begin{equation}
        N(P,x) \geq [\Pi_{i=1}^k N(P,x^*_i)]\cdot exp(\Delta\cdot \sum_{i=1}^k L_i)
    \end{equation}
    
    Where $x^*_1...x^*_k$ is the lowest generation complexity decomposition in $d(x)$ and $L_1...L_k$ is its minimal solution length.
\end{theorem}
This bounds the success rate of solving a general problem in terms of simpler tasks that the model may choose to decompose the problem to. Proof presented in appendix \ref{Proof:theorem_2}.

\section{Experiments}
In this subsection we test the assumptions and results of our theory. We create simple compositional coding problems with pairs of problems from the Human Eval benchmark \citep{chen2021codex} and code contests dataset \citep{doi:10.1126/science.abq1158}. The choice to use pairs of problems and not more, is the low pass rate of $\sim 1/200$ of composition, requiring many samples.
First, we show the results of theorem \ref{theorem:1}, stating that composition is typically harder than solving the the subproblems in parallel. Then, we show explicit indications for the exponential length dependence of compositional hardness, by comparing probabilities of solutions with/without composition, as theoretically suggested in lemma \ref{lemma:1}. Finally, we look at our assumption \ref{assumption:1} on the noise inserted into the logits of the second problem, and observe that it is indeed large enough to interfere with the decoding process (assumption \ref{assumption:1}). The experiments were performed on Llama-3-8B-Instruct \citep{dubey2024llama}. In appendix \ref{sec:mistral} we show the results apply also to Mistral-7B-Instruct-v0.3 \citep{jiang2023mistral}, and in appendix \ref{sec:70B}, we present results for Llama-3-70B-Instruct, to test dependence on model size. For additional experimental details, see appendix \ref{experimental_details}.

\subsection{Generation Complexity Results}\label{generation_complexity_experiment}
Here we test the actual generation complexity of an LLM to different problems corresponding to scenarios from our theoretical results.
As proposed in the theoretical section, the generation complexity to solve two problems in a single context can be much higher than that of parallel contexts if the model was not explicitly trained on such a task. To test this, we built a set of composite problems based on the Human Eval benchmark \citep{chen2021codex} and code contests dataset \citep{doi:10.1126/science.abq1158}.

To create composite problems, we took pairs of problems from Human Eval and code contests, and created from each a problem whose solution requires to explicitly solve both problems. Additionally, for harder problems, we created compositions of problems from code contests dataset.
We used the following two main templates (and an additional template for human eval presented in appendix \ref{experimental_details}):

\begin{itemize}
    \item Human Eval -- Problem 1 and Problem 2 have an integer output. Composition is to solve both and print the product of their outputs.

        \begin{figure}[h!]
        \centering
        \includegraphics[width=0.6\linewidth]{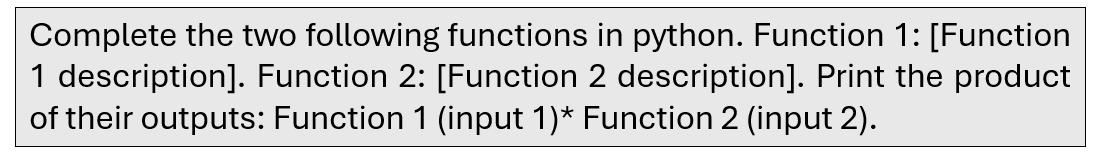}
        \label{fig:template_1}
    \end{figure}

    \item Code Contests -- Problem 1 and problem 2 are problems from the dataset. Composition is to read the inputs sequentially and print the outputs sequentially.
        \begin{figure}[h!]
        \centering
        \includegraphics[width=0.6\linewidth]{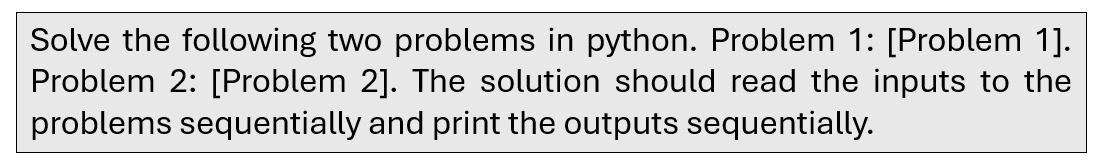}
        \label{fig:template_3}
    \end{figure}
\end{itemize}
We denote the composite problem of $x_1$ and $x_2$ as $x_1\oplus x_2$. While these compositions are synthetic, their purpose is to demonstrate that the LLM suffers from a ``mental load" when asked to solve multiple tasks within the same context, similarly to a problem decomposition obtained from a CoT, which leads to the result of theorem \ref{theorem:1}.
For each problem we sampled 200 solutions and evaluated the generation complexity as the inverse of percentage of correct solutions.

\begin{figure}[h!]
    \centering
    \includegraphics[width=0.9\linewidth]{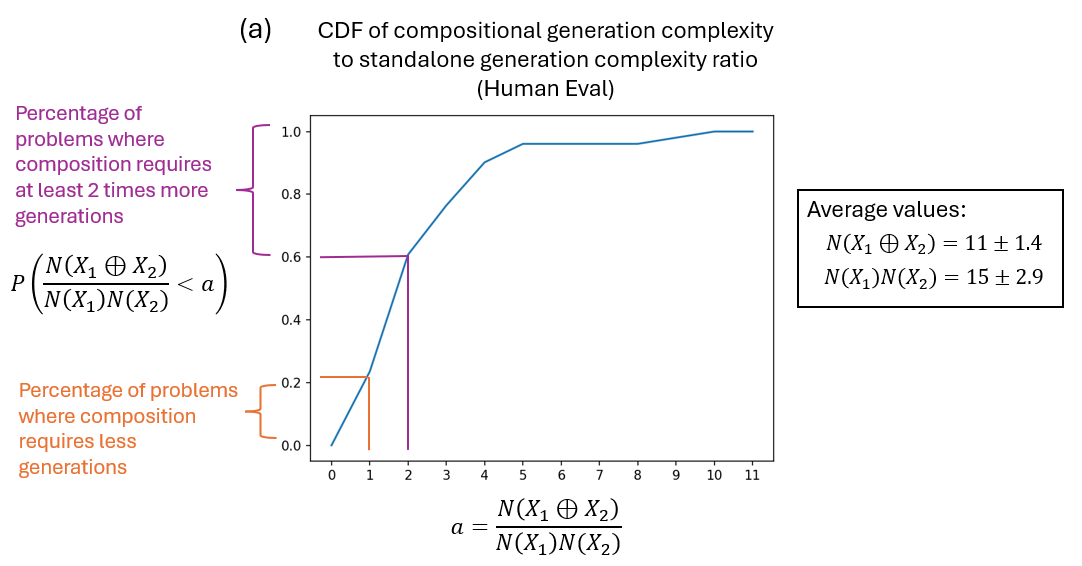}
    \includegraphics[width=0.9\linewidth]{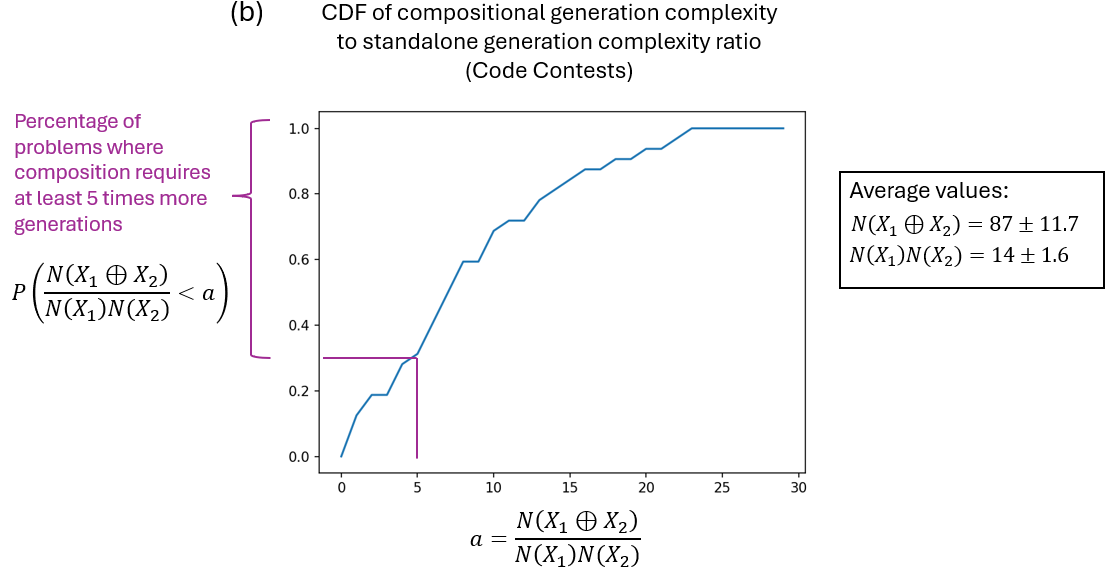}

    \caption{Cumulative distribution function for the ratio of generation complexity using composition, $N(P,x)$, to product of generation complexities for the standalone problems, $N(P,x_1)\cdot N(P,x_2)$ (corresponding to the multi-agent generation complexity). The x axis denotes values for the ratio of generation complexities, the y axis is the percentage of problems in which the ratio is no larger than this value (\eg~for $a=5$, the y axis value is the percentage of problems where composition requires up to $\times 5$ more samples than the multi-agent case). (a) For the human eval composition. As seen, in most cases composition requires twice more samples, and sometimes up to 10 times more samples. (b) For the code contests composition. As seen, the majority of problems have a factor of at least 5, and some up to 20.}
    \label{fig:cdf}

\end{figure}

Figure \ref{fig:cdf} shows the cumulative distribution function (CDF) of the compositional hardness -- the value along the $y$ axis is the CDF of $\frac{N(P,x)}{N(P,x_1)N(P,x_2)}$, \ie~  percentage of compositional problems, in which $\frac{N(P,x)}{N(P,x_1)N(P,x_2)}$ is smaller than a given value. \eg, $x=5$, the value on the $y$ axis, is percentage of problems in which composition requires up to $\times 5$ more generations relative to solving independently. As can be seen, for most of the problems, composition generation complexity is larger than the product of generation complexities of the components (up to factors of $10-20$).

As seen in theorem \ref{theorem:1}, we have:

\begin{equation}
    N(P,x_1\oplus x_2) \gg N(P,x_1)N(P,x_2)
\end{equation}
Furthermore, in code contests, where the problems typically have longer solutions (few hundred tokens) and are harder, thus the model is less certain, the compositional hardness is much larger. If one only has access to an end-to-end verifier of the composite problem, it is more advantageous to generate independently for each problem then to generate for the composite problem, giving the advantage to a multi-agent system over a single LLM instance.

Similar results are shown for Mistral-7B-Instruct-v0.3 in appendix \ref{sec:mistral} In appendix \ref{sec:70B}, we present results for composition on Llama-3-70B-Instruct, and find that on the same code contests compositions, the compositional generation complexity improves significantly relative to its 8B counterpart.
However, with slightly harder compositions of code contest problems involving four problems instead of two, the 70B model's performance on composition drops significantly, with most compositions requiring over 5-10 times more generations than the non-compositional case. This hints larger models are more efficient at composition yet still suffer from this effect as compositional difficulty rises.

\subsection{Exponential Length Dependence of Compositional Hardness}\label{exponential_length_experiment}
Here, we used the same compositions as in the previous subsection, and measured the difference in probabilities of correct solutions with vs without composition as a function of the number of tokens in the solution. 
``Correct" solutions were taken as the canonical solutions of the dataset, as they are ``neutral" in the sense that solutions generated by the model with/without composition may have different styles, and measuring the probability of these generated sequences may create an artificial bias in favor of one of the two.

\begin{figure}[h!]
    \centering
    \includegraphics[width=0.9\linewidth]{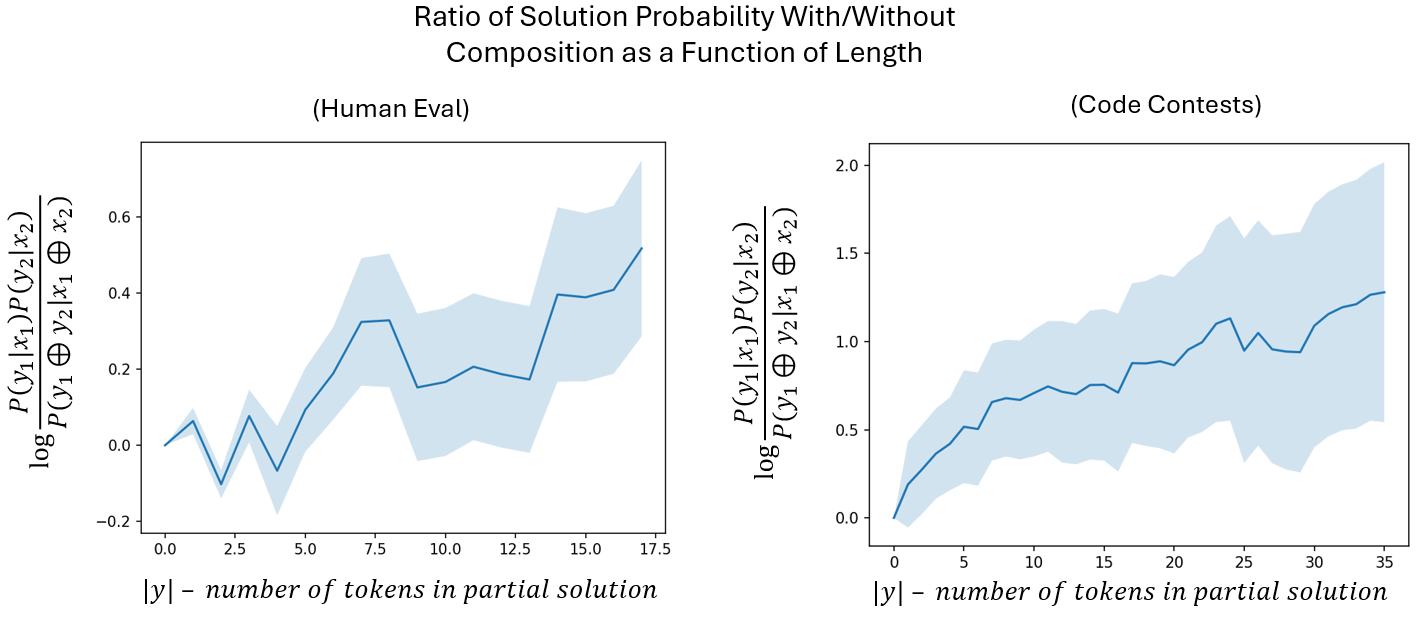}
    
    \caption{Ratio of correct solution probability with vs without composition. An exponential trend is \wrt function length is observed.}
\label{fig:exponential_length}
\end{figure}

As suggested by lemma \ref{lemma:1}, we expect to see on average:
\begin{equation}
   \log\frac{P(y_1|x_1)P(y_2|x_2)}{P(y_1\oplus y_2|x)} \geq \frac{\Delta}{4}\cdot(|y_1|+|y_2|)  
\end{equation}
Meaning that the log ratio of correct solutions without/with composition increases linearly with length.
As can be seen in figure \ref{fig:exponential_length}, an exponential increase in probability of the correct solution is observed without composition relative to with composition, as a function of length of the code. 
As the y axis is plotted in the log domain, the linear curve approximates $\Delta/4$, taking a value of $\approx 0.05$. This means a decrease of $e^{-1}$ in success rate for every 20 tokens, matching the typical increase in generation complexity seen in figure \ref{fig:cdf}. In practice, $\Delta$ is likely larger for some compositions, as the average is over many sequences, yet fluctuations are high.

While measuring probability of sequences not generated by the model is less reliable than the above experiment in \ref{generation_complexity_experiment}, which shows the gap of compositional hardness in a real sampling scenario, this approach is useful to qualitatively show an exponential length dependence for sequences in compositional scenarios vs non-compositional, which is expected to hold during generation.

\subsection{Experiments on Assumptions}\label{exp:assumptions}
Here we look at our assumption of noise inserted into the logits (assumption \ref{assumption:1}).

\begin{figure}[h!]
    \centering
    \includegraphics[width=0.9\linewidth]{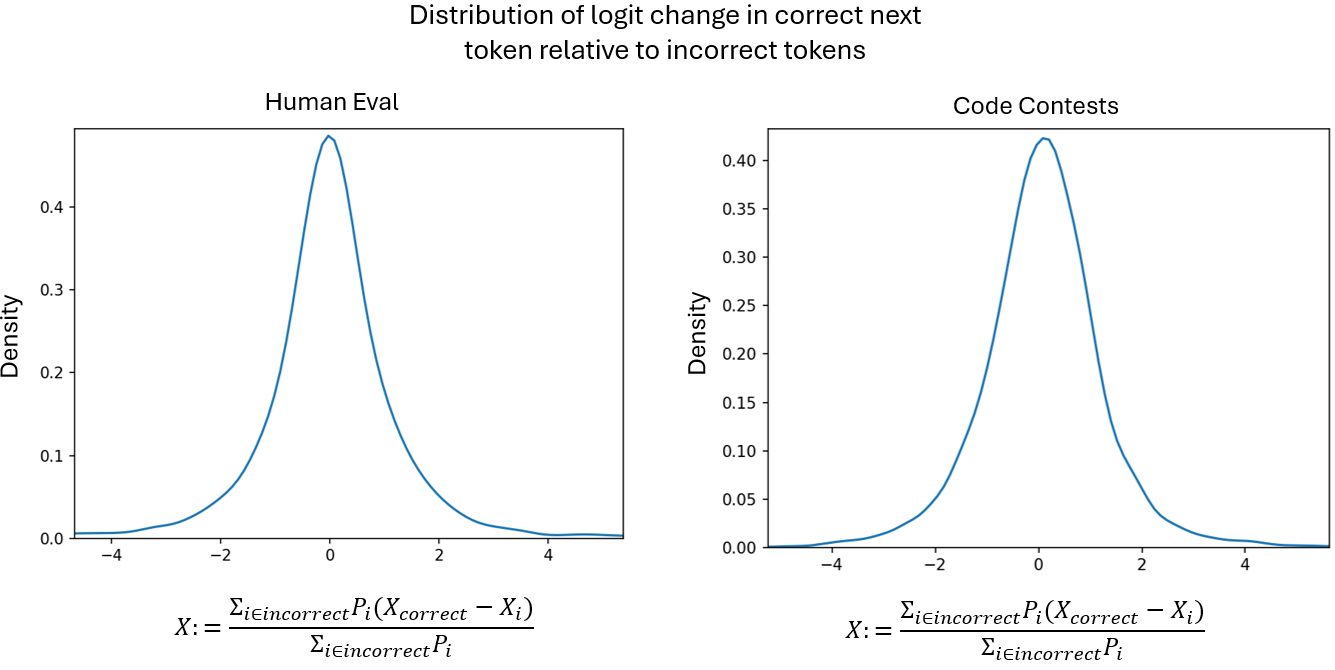}
    \caption{Change in logits of correct tokens minus incorrect tokens due to composition.}
\label{fig:logit_noise}
\end{figure}

\paragraph{Noise distribution: }Using the same compositions as before, we measure the change to the logits of correct tokens with vs without composition. As in the theoretical assumption, we subtract this with the mean change in the logits of the incorrect tokens: 
\begin{equation}
    X=\frac{\sum_{i\in incorrect}P_i \langle noise, U^Te_i-U^Te_{correct}\rangle}{\sum_{i\in incorrect}P_i}
\end{equation}
We calculate this over different sequences. As can be seen in figure \ref{fig:logit_noise}, the change in logits creates a noise that is symmetric, bounded, $X<M\approx 4$, and has finite absolute deviation $E[|X|]>0$, in accordance with assumption \ref{assumption:1}.

\paragraph{Estimation of $\Delta(\epsilon,X)$ and $\sigma(\epsilon,X)$:} A ballpark estimation to the theoretical constants based on the estimated noise is presented in appendix \ref{sec:delta_sigma_estimation}, that matches the estimation of $\Delta$ in the range of $0.05$ to $0.2$ from the above subsection. Similarly, we estimate $\sigma(\epsilon,X)$, with typical values of $\sigma\approx 1- 2$.

\section{Discussion}
In this work, we point a limitation of LLMs' ability to perform several sub-tasks within the same context window -- an in-context hardness of composition, pointing to an advantage for distributing a decomposed problem in a multi-agent system of LLMs over using a single LLM instance. The hardness of composition is quantified by the generation complexity metric, \ie, the number of LLM generations required to sample at least one correct solution. We found an exponential gap between generation complexity of solving a composition problem within the same context relative to distributing it among multiple agents. This is attributed to the transformer's nature of using all tokens in the context simultaneously for decoding, which inserts noise into generated sequences, caused from mixing in sub-tasks' latent representations (screening). Consequently, even if a model has the ability to perform two tasks, it may not be able to perform them both within the same context, in agreement with empirical work of \cite{zhuo2024bigcodebench}, showing SOTA LLMs still perform poorly on complex coding tasks.
This provides a view of the model's effective context length through the lens of screening, which is the model's ability to isolate the relevant context at each decoding step -- additional irrelevant context may reduce model's performance on other tasks within the context window exponentially with length. 
Lastly, we point that while advantegous, the use of multi-agent systems may introduce new challenges, such as coherence between agents, which we leave for future work.

\section*{Acknowledgments}
We thank Yoav Levine and Noam Wies for fruitful discussions. This research was supported by the ERC (European Research Council) and the ISF (Israel Science Foundation).

\bibliography{main}
\bibliographystyle{arxiv_upload}

\clearpage

\appendix

\section{Proof of Lemma \ref{lemma:1}}\label{proof_lemma}
Here we prove lemma \ref{lemma:1}. We formulate the detailed lemma:
\begin{lemma}\label{lemma:1_detailed}
Let $\epsilon,\delta\in(0,1)$, and $M> 0$. Let $x$ be a compositional problem and $y_1 \oplus y_2$ a solution, with $x_1$, $x_2$ being the corresponding sub-problems. Suppose that the noise injected to the logits as defined in equation \ref{eq:weighted_noise}, satisfies assumption \ref{assumption:1} - for all decoding steps, it is continuous, symmetric and bounded within $[-M,+M]$. Suppose further that the probability assigned to the correct token at each decoding step is bounded within $[\epsilon,1-\epsilon]$. Denote by $\Delta:=\expectation_X[\log(\epsilon+(1-\epsilon)e^X)]$ and $\sigma^2:=Var_X[\log(\epsilon+(1-\epsilon)e^X)]$ the renormalizing term's mean and variance (as defined in equations \ref{eq:delta} and \ref{eq:sigma} respectively). Under the assumption, $\Delta,\sigma$ are strictly positive, and
if $|y_1|+|y_2|>\frac{M^2}{\sigma^2\cdot h(\frac{3\Delta\cdot M}{4\sigma^2})}\ln\frac{1}{\delta}$, where $h(x)=(x+1)\ln(1+x)-x>0$, we have with
probability of at least $1-\delta$ that:
\begin{equation}
    P(y_1\oplus y_2|x)\leq P(y_1|x_1)\cdot P(y_2|x_2)e^{-\frac{\Delta \cdot(|y_1|+|y_2|)}{4}}
\end{equation}
Where $P(y|x)$ is the probability assigned to $y$ by the model, given context $x$.
\end{lemma}
Thus $f$ from the main text is $f(\sigma,M,\Delta)=\frac{M^2}{\sigma^2\cdot h(\frac{3\Delta M}{4\sigma^2})}$.

\textit{Proof:}

We start by proving the following lemma:

\begin{lemma}
Let $p\in(0,1)$ and $X$ a random variable over real numbers. Denote by:
$\Delta\left(p,X\right)=E\left[\log{\left(p+\left(1-p\right)e^X\right)}\right]$.
Claim: For a continuous symmetric random variable, X, we have $\Delta\left(p,X\right)>0$. Furthermore, if $p\in(\epsilon,1-\epsilon)$, then $\Delta(p,X)>\Delta(\epsilon,X)=\Delta(1-\epsilon,X)$
\end{lemma}

\textit{Proof:} 

Denote by $\rho(X)$ the density function of $X$, then:
\begin{equation}
    E\left[\log{\left(p+\left(1-p\right)e^{X}\right)}\right]=\int_{-\infty}^{0}{\log{\left(p+\left(1-p\right)e^x\right)}\rho(x)}+\int_{0}^{\infty}{\log{\left(p+\left(1-p\right)e^x\right)}\rho(x)}=
\end{equation}
Switching sign of the integration variable in the second term:
\begin{equation}
    =\int_{0}^{\infty}{\log{\left(p+\left(1-p\right)e^{-x}\right)}\rho(-x)}+\int_{0}^{\infty}{\log{\left(p+\left(1-p\right)e^x\right)}\rho(x)}=
\end{equation}
Using the symmetry condition on $X$:
\begin{equation}
    =\int_{0}^{\infty}{\log{\left(p+\left(1-p\right)e^{-x}\right)}\rho(x)}+\int_{0}^{\infty}{\log{\left(p+\left(1-p\right)e^x\right)}\rho(x)}=
\end{equation}
\begin{equation}
    =\int_{0}^{\infty}{\left(\log{\left(p+\left(1-p\right)e^{-x}\right)}+\log{\left(p+\left(1-p\right)e^x\right)}\right)\rho(x)}=
\end{equation}
Again, applying the symmetry of $X$:
\begin{equation}
    =\frac{1}{2}\int_{-\infty}^{\infty}\left(\log{\left(p+\left(1-p\right)e^{-x}\right)}+\log{\left(p+\left(1-p\right)e^x\right)}\right)\rho\left(x\right)
\end{equation}
The integrand is positive:
\begin{equation}
    \mathrm{log}(p+(1-p)e^x)+\mathrm{log}(p+(1-p)e^{-x})\mathrm{=}
\end{equation}
\begin{equation}
    =\log{(p^2+\left(1-p\right)^2+p\left(1-p\right)\left(e^x+e^{-x}\right)}=
\end{equation}
\begin{equation}
    \log{(1+}p\left(1-p\right)\left(e^x+e^{-x}-2\right))\ 
\end{equation}
For $x=0$, the argument inside the log equals $1$, otherwise, the argument inside the log is always larger than $1$, since $e^x+e^{-x}-2>0$.
Hence the integrand is positive everywhere except in $x=0$, meaning the integral is positive for any continuous symmetric $X$.
Thus, for any $p$ the expectation value is positive.

\begin{equation}
    E\left[\log{\left(p+\left(1-p\right)e^{X}\right)}\right]=\Delta\left(p,X\right)>0
\end{equation}
Furthermore it is symmetric around $p=1/2$, and monotonic for $p\rightarrow0$ and $p\rightarrow1$:
\begin{equation}
    \mathrm{log}(p+(1-p)e^x)+\mathrm{log}(p+(1-p)e^{-x})\mathrm{=}
\end{equation}
\begin{equation}
    =\log{(p^2+\left(1-p\right)^2+p\left(1-p\right)\left(e^x+e^{-x}\right)}=
\end{equation}
\begin{equation}
    \log{(1+}p\left(1-p\right)\left(e^x+e^{-x}-2\right))\ 
\end{equation}
As can be seen, it is monotonic with $p(1-p)$, and if $p\in(\epsilon,1-\epsilon)$, it is minimal for $p=\epsilon,1-\epsilon$.

\paragraph{Proof of Main Lemma:}

We now move to the proof of the main lemma. We show the exponential growth with the length of the solution to the second problem, $y_2$ (the proof for the exponential dependence on $y_1$ is identical). Let us look at the probability that the model assigns the $n$'th token in the sequence $y_2$ in the compositional problem. It is given by the softmax on the projections of the final hidden layer representation on the vocabulary $V$ given the context:
\begin{equation}
    P\left(y_2\left[n\right]\middle| x\oplus y_1\oplus y_2\left[:n\right]\right)=\frac{e^{\left\langle r^{(L)}\left(x\oplus y_1\oplus y_2\left[:n\right]\right),U^Te_{y_2\left[n\right]}\right\rangle}}{e^{\left\langle r^{(L)}\left(x\oplus y_1\oplus y_2\left[:n\right]\right),U^Te_{y_2\left[n\right]}\right\rangle}+\Sigma_{i\in\left[V\right]\backslash\{y_2[n]\}}e^{\left\langle r^{(L)}\left(x\oplus y_1\oplus y_2\left[:n\right]\right),U^Te_i\right\rangle}}
\end{equation}

According to assumption 1, $\left\langle r^{(L)}\left(x\oplus y_1\oplus y_2\left[:n\right]\right),U^Te_i\right\rangle=\left\langle r^{(L)}\left(x_2\oplus y_2\left[:n\right]\right),U^Te_i\right\rangle+X_i$, meaning the model is receiving a noisy version of the representation to the problem it is trying to solve, where $X_i$ is the noise onto the $i$'th token.

\begin{equation}\label{eq:noisy_logits}
    =\frac{P\left(y_2\left[n\right]\middle| x_2\oplus y_2\left[:n\right]\right)e^{X_{y_2\left[n\right]}}}{P\left(y_2\left[n\right]\middle| x_2\oplus y_2\left[:n\right]\right)e^{X_{y_2\left[n\right]}}+\Sigma_{i\in\left[V\right]\backslash\{y_2[n]\}\}}P\left(i\middle| x_2\oplus y_2\left[:n\right]\right)e^{X_i}}
\end{equation}

For brevity, denote $P_0=P\left(y_2\left[n\right]\middle| x_2\oplus y_2\left[:n\right]\right)$, and $P_i=P\left(i\middle| x_2\oplus y_2\left[:n\right]\right)$.

\begin{equation}
    =\frac{P_0e^{X_0}}{P_0e^{X_0}+\Sigma_{i\in\left[V\right]\backslash\{0\}}P_ie^{X_i}}
\end{equation}

Now, using the Jensen's inequality in the denominator:

\begin{equation}
    \le\frac{P_0e^{X_0}}{P_0e^{X_0}+\left(1-P_0\right)e^{\Sigma_{i\in\left[V\right]\backslash\{0\}}\frac{P_iX_i}{1-P_0}}}=\frac{P_0}{P_0+\left(1-P_0\right)e^{\Sigma_{i\in\left[V\right]\backslash\{0\}}\frac{P_iX_i}{1-P_0}-X_0}}
\end{equation}

Now, denote $X=\Sigma_{i\in\left[V\right]\backslash\{0\}}\frac{P_i(X_i-X_0)}{1-P_0}$, according to assumption \ref{assumption:1}, it is a symmetric, continuous random variable, bounded between $[-M,+M]$. Rewriting the above as:
\begin{equation}\label{eq:noise_logit_bound}
    = P_0 e^{-\log(P_0 +(1-P_0)e^X)}
\end{equation}
So the correct token probability with composition $P_0'$ is decreased by the factor in the exponent $\rightarrow P_0 e^{\log(P_0 +(1-P_0)e^X)}$, relative to the probability without composition, $P_0$. For a full sequence, we apply the probability chain rule and obtain the following:

\begin{equation}\label{eq:sequence_bound}
    P\left(y_2\middle| x\oplus y_1\right)=P\left(y_2\right|x_2)e^{-\Sigma_{i=1}^{\left|y_2\right|}\log{\left(P_0^i+\left(1-P_0^i\right)e^{X_i}\right)}}
\end{equation}
Where $P_0^i$ is the probability of the correct token in the $i$'th step without composition.
Now, because $E_{X_i}\left[\log{\left(P^i_0+\left(1-P^i_0\right)e^{X_i}\right)}\right]=\Delta\left(P^i_0,X_i\right)>0$, from the above lemma, we get a sum of random variables with mean that is larger than zero. We will use a concentration inequality to bound it.

We start with bounding the random variables:
\begin{equation}
    \log{\left(P^i_0+\left(1-P^i_0\right)e^{X_i}\right)}<\log{\left(P^i_0+\left(1-P^i_0\right)e^M\right)}\le M
\end{equation}
\begin{equation}
    \log{\left(P^i_0+\left(1-P^i_0\right)e^{X_i}\right)}>\log{\left(P^i_0+\left(1-P^i_0\right)e^{-M}\right)}\geq-M
\end{equation}

Also notice that since $P_0\in(\epsilon,1-\epsilon)$, the above lemma implies:
\begin{equation}
    \Delta\left(P_0^i,X_i\right)>\Delta\left(\epsilon,X_i\right)
\end{equation}

Thus from linearity of the expectation value of the sum $S=\Sigma_{i=1}^{\left|y_2\right|}\log{\left(P_0+\left(1-P_0\right)e^{X_i}\right)}$ is:

\begin{equation}\label{eq:expectation_bound}
    E[S]=E\left[\Sigma_{i=1}^{\left|y_2\right|}\log{\left(P_0+\left(1-P_0\right)e^{X_i}\right)}\right]>\left|y_2\right|\cdot\Delta(\epsilon,X)
\end{equation}
Similarly, if we denote $\sigma^2(\epsilon,X):=Var_X[\log (\epsilon+(1-\epsilon)e^X)]]$
(which is no larger than $M$), we can upper bound the variance of the sum:
\begin{equation}
    Var[S]=Var\left[\Sigma_{i=1}^{\left|y_2\right|}\log{\left(P_0+\left(1-P_0\right)e^{X_i}\right)}\right]<\left|y_2\right|\cdot\sigma(\epsilon,X)^2
\end{equation}
This is because $Var_X[\log (p+(1-p)e^X)]] \leq Var_X[\log (\epsilon+(1-\epsilon)e^X)]]$ (see proof in appendix \ref{lemma:variance_bound})

We can then apply Bennet's inequality:
\begin{equation}
P\left(S-\expectation[S] < -t\right) \leq \exp{\left(-\frac{Var[S]}{M^2}h{\left(\frac{t M}{Var[S]}\right)}\right)}
\end{equation}
Where $h(x)=(1+x)\log(1+x)-x$ and $M$ is the bound for each summand in $S$.
Next, we take $t=\expectation[S]-\frac{\Delta(\epsilon,X)}{4}|y_2|$. Plugging this in the above yields:
\begin{equation}
P\left(S < \frac{\Delta(\epsilon,X)}{4}|y_2|\right) \leq \exp{\left(-\frac{Var[S]}{M^2}h{\left(\frac{t M}{Var[S]}\right)}\right)}
\end{equation}
We note that $t=\expectation[S]-\frac{\Delta(\epsilon,X)}{4}|y_2| > \frac{3\Delta(\epsilon,X)}{4}|y_2|$ (from equation \ref{eq:expectation_bound}). Thus due to the (increasing) monotonicity of $h(\frac{tM}{Var[S]})$ \wrt~$t$ (hence decreasing monotonicity in the exponent), we have:
\begin{equation}
P\left(S < \frac{\Delta(\epsilon,X)}{4}|y_2|\right) \leq \exp{\left(-\frac{Var[S]}{M^2}h{\left(\frac{3\cdot\Delta(\epsilon,X)|y_2| M}{4\cdot Var[S]}\right)}\right)}
\end{equation}
Due to the (increasing) monotonicity of the exponent's argument in $Var[S]$ which is upper bounded by $|y_2|\sigma^2$, we get:
\begin{equation}
P\left(S < \frac{\Delta(\epsilon,X)}{4}|y_2|\right) \leq \exp{\left(-\frac{|y_2|\sigma^2}{M^2}h{\left(\frac{3\cdot\Delta(\epsilon,X) M}{4\cdot \sigma^2}\right)}\right)}
\end{equation}

Looking at the complementary event to the one in the above equation ($S\geq \frac{\Delta(\epsilon,X)}{4}|y_2|$), and plugging in the definition for $S$, we get:
\begin{equation}\label{eq:concentration_inequality}
    P\left(\Sigma_{i=1}^{\left|y_2\right|}\log{\left(P_0^i+\left(1-P_0^i\right)e^{X_i}\right)}>\frac{1}{4}\left|y_2\right|\cdot\Delta(\epsilon,X)\right)\geq 1-\exp{\left(-\frac{|y_2|\sigma^2}{M^2}h{\left(\frac{3\Delta M}{4\sigma^2}\right)}\right)}
\end{equation}

Let $\delta>0$, then for:
\begin{equation}
    \left|y_2\right|>\frac{M^2}{\sigma^2 h(\frac{3\Delta M}{4\sigma^2})}\cdot\ln{\frac{1}{\delta}}
\end{equation}
We obtain from equation \ref{eq:concentration_inequality} with probability of at least $1-\delta$ that $\Sigma_{i=1}^{\left|y_2\right|}\log{\left(P_0^i+\left(1-P_0^i\right)e^{X_i}\right)}>\frac{1}{4}\left|y_2\right|\cdot\Delta(\epsilon,X)$. Plugging this back into equation \ref{eq:sequence_bound}:
\begin{equation}
    P\left(y_2\middle| x\oplus y_1\right)<P\left(y_2\right|x_2)e^{-\frac{\Delta}{4}|y_2|}
\end{equation}
With probability $1-\delta$.

Using the same idea for $y_1$, we obtain:
\begin{equation}
    P\left(y_1\middle| x\right)<P\left(y_1\right|x_1)e^{-\frac{\Delta}{4}|y_1|}
\end{equation}

Thus together, if $    \left|y_1\right|+\left|y_2\right|>\frac{M^2}{\sigma^2 h(\frac{3\Delta M}{4\sigma^2})}\cdot\ln{\frac{1}{\delta}}$, we have with probability $1-\delta$ that:
\begin{equation}
    P\left(y_1\oplus y_2\middle| x\right) < P\left(y_1\right|x_1)P\left(y_2\right|x_2)e^{-\frac{\Delta}{4}(|y_1|+|y_2|)}
\end{equation}

\section{Proof of theorem \ref{theorem:1}}\label{proof_theorem}
We state theorem \ref{theorem:1} with the full details:
\begin{theorem}\label{theorem:1_detailed}
    Let $\epsilon,\delta\in (0,1)$, and $N,M>0$. Let $x$ be a compositional problem, with $x_1$, $x_2$ being the corresponding sub-problems. Denote by $L_1, L_2$ the minimal solution length to $x_1, x_2$ respectively, and the total number of solutions to $x$ by $N$. Define $\Delta,\sigma$, the renormalizing term's mean and variance (as defined in equations \ref{eq:delta} and \ref{eq:sigma} respectively) and by $M$ the bound on the logit noise (assumption \ref{assumption:1}). Under the assumptions of lemma \ref{lemma:1}, they are strictly positive, $\Delta,\sigma,M>0$, and if the minimal solution length $L_1+L_2$, satisfies $L_1+L_2>\frac{M^2}{\sigma^2\cdot h(\frac{3\Delta\cdot M}{4\sigma^2})}\ln\frac{N}{\delta}$, where $h(x)=(x+1)\ln(1+x)-x>0$, we have with probability of at least $1-\delta$ that the generation complexity (definition \ref{def:gen_comp}) satisfies:
    \begin{equation}
        N(P,x)\geq N(P,x_1)N(P,x_2)\cdot e^{\frac{\Delta \cdot (L_1+L_2)}{4}}
    \end{equation}
\end{theorem}

\textit{Proof:}

Now, suppose there are $N$ solutions to the problem, all of length $\geq L$ (larger than the minimal description length), then we need to use a union bound. We get the result of the lemma \ref{lemma:1} over all sequences with probability:
\begin{equation}
    \left(1-\exp{\left(-\frac{L\sigma^2}{M^2}h{\left(\frac{3\Delta M}{4\sigma^2}\right)}\right)}\right)^N\geq1-N\exp{\left(-\frac{L\sigma^2}{M^2}h{\left(\frac{3\Delta M}{4\sigma^2}\right)}\right)}
\end{equation}
Require this to equal:
\begin{equation}
    =1-\delta
\end{equation}
Thus for:
\begin{equation}  \left|y_1\right|+\left|y_2\right|>\frac{M^2}{\sigma^2 h(\frac{3\Delta M}{4\sigma^2})}\cdot\ln{\frac{N}{\delta}}
\end{equation}
We obtain with probability of at least $1-\delta$ that all solutions satisfy the result of lemma \ref{lemma:1}. If the minimal solution length is $L_1+L_2$, for all solutions to satisfy the inequality, we require:
\begin{equation}  
L_1+L_2>\frac{M^2}{\sigma^2 h(\frac{3\Delta M}{4\sigma^2})}\cdot\ln{\frac{N}{\delta}}
\end{equation}

Thus we have with probability $1-\delta$ that:
\begin{equation}
    N(P,x) = \frac{1}{\sum_{y_1,y_2\in\textit{correct solutions}}P(y_1\oplus y_2|x)} \geq ~~~~~~~~~~~~~~~~~~~~~~~~~~~~~
\end{equation}
\begin{equation}
   ~~~~~~~~~~ \geq  \frac{1}{\sum_{y_1,y_2\in\textit{correct solutions}}P(y_1|x_1)P(y_2|x_2)}e^{\frac{\Delta}{4}(L_1+L_2)}=
\end{equation}

\begin{equation}
    =\frac{1}{\sum_{y_1\in\textit{correct solutions for $x_1$}}P(y_1|x_1)\sum_{y_1\in\textit{correct solutions for $x_2$}}P(y_2|x_2)}e^{\frac{\Delta}{4}(L_1+L_2)}
\end{equation}
\begin{equation}
    =N(P,x_1)N(P,x_2)e^{\frac{\Delta}{4}(L_1+L_2)}
\end{equation}

\section{Proof of Variance Bound}\label{lemma:variance_bound}
Here we show that the variance of the noise $\sigma^2(p,X)$ is maximal for $p=\epsilon$.
\begin{lemma}
    For $p\in(\epsilon,1-\epsilon)$ and a random variable $X$, it holds that:
    \begin{equation}
        Var_X[\log(p+(1-p)e^X)] \leq Var_X[\log(\epsilon+(1-\epsilon)e^X)]
    \end{equation}
\end{lemma}

\textit{proof:}

Consider the transformation $T_p(x)=\log(p+(1-p)e^x)$, notice that its derivative \wrt $x$ is:
\begin{equation}
    \frac{dT_p}{dx}=\frac{(1-p)e^x}{p+(1-p)e^x}
\end{equation}
Which always takes values in $(0,1)$. Thus due to strict monotonicity, it is an invertible map for any $p\in(\epsilon,1-\epsilon)$. Additionally, for any $x$, we have:
\begin{equation}
    \frac{dT_p}{dx} =\frac{(1-p)e^x}{p+(1-p)e^x}\leq \frac{(1-\epsilon)e^x}{\epsilon+(1-\epsilon)e^x}= \frac{dT_\epsilon}{dx}
\end{equation}

Next, we look at:
\begin{equation}
    Var_X[T_p(x)] = Var_X[T_p(T_\epsilon^{(-1)}T_\epsilon)(x)]=Var_X[(T_p\circ T_\epsilon^{(-1)})(T_\epsilon(x))] 
\end{equation}

Now, for an $M$-Lipschitz map, $T$, we have $Var_X[T(X)] \leq M^2 Var[X]$, therefore:
\begin{equation}
   Var_X[T_p(x)] \leq \sup_{x}|\frac{d(T_p\circ T_\epsilon^{(-1)})}{dx}|^2 Var_X[T_\epsilon(x)]
\end{equation}

Since the map $T_p\circ T_\epsilon^{(-1)}$, has a derivative that is bounded by:
\begin{equation}
    |\frac{d(T_p\circ T_\epsilon^{(-1)})}{dx}| \leq \frac{|\frac{d(T_p)}{dx}|}{|\frac{d(T_\epsilon)}{dx}|} \leq 1
\end{equation}
We obtain:
\begin{equation}
   Var_X[T_p(x)] \leq  Var_X[T_\epsilon(x)]
\end{equation}
Plugging in the definition of $T_p$, we get:
    \begin{equation}
        Var_X[\log(p+(1-p)e^X)] \leq Var_X[\log(\epsilon+(1-\epsilon)e^X)]
    \end{equation}
As desired.

\section{Proof of theorem \ref{theorem_2}}\label{Proof:theorem_2}
As explained in section \ref{sec:gen_comp}, the probability that the model provides a correct final answer is the probability to generate a correct COT decomposition $x_1...x_k\in d(x)$ (where $d(x)$ denotes the valid decompositions to $x$ for which the model asigns non-zero probability $P(x_1...x_k|x)>0$), followed by a correct implementation, $y_1\oplus...\oplus y_k$, where for each task, the set of correct answers is denoted by $s(x_i)$:
\begin{equation}
    P[\textit{generate correct solution to x}]  =
\end{equation}
\begin{equation}
    =\sum_{(x_1...x_k)\in d(x)}\sum_{y_1\oplus...\oplus y_k:\forall i\in[k]\textit{ } y_i\in s(x_i)}P(x_1...x_k\oplus y_1\oplus...\oplus y_k|x)
\end{equation}
Applying the probability chain rule we can write it as:
\begin{equation}
    =\sum_{(x_1...x_k)\in d(x)}P(x_1...x_k|x)\sum_{y_1\oplus...\oplus y_k:\forall i\in[k]\textit{ } y_i\in s(x_i)}P( y_1\oplus...\oplus y_k|x,x_1...x_k)
\end{equation}
The inner sum is the generation complexity of the compositional problem:
\begin{equation}
    =\sum_{(x_1...x_k)\in d(x)}\frac{P(x_1...x_k|x)}{N(P,(x,x_1...x_k))}
\end{equation}
Since the generation complexity to the problem is the inverse success rate of solving it, we obtain:
\begin{equation}
    \frac{1}{N(P,x)}=\sum_{(x_1...x_k)\in d(x)}\frac{P(x_1...x_k|x)}{N(P,(x,x_1...x_k))}
\end{equation}
Essentially, the success rate of solving the problem is a weighted sum over the valid decompositions and the success rate of solving them.
Since this is a weighted sum, and the sum of weights does not surpass one, we have:
\begin{equation}
    \frac{1}{N(P,x)} \leq \max_{(x_1...x_k)\in d(x)}\frac{1}{N(P,(x,x_1...x_k))}
\end{equation}
Meaning the easiest decomposition for the model to solve is a bound. Assuming at the very list this has finite probability (else the generation complexity is infinite and there is nothing to prove), we can invert the equation:
\begin{equation}
    N(P,x) \geq \min_{(x_1...x_k)\in d(x)}N(P,(x,x_1...x_k))
\end{equation}

So we are left with showing that the results of theorem \ref{theorem:1} hold for $\min_{(x_1...x_k)\in d(x)}N(P,(x,x_1...x_k))$. This requires a union bound over $d(x)$. Essentially we need to replace the $\delta$ from that theorem to $\frac{\delta}{|d(x)|}$, so we obtain:
\begin{equation}
    \sum_{i=1}^k L_i \geq \frac{M^2}{\sigma^2 h(\frac{3\Delta M}{4\sigma^2})}\cdot \ln\frac{N\cdot|d(x)|}{\delta}
\end{equation}
As desired.

As for why theorem \ref{theorem:1} holds for more than a pair of problems can be proven by revising the proof of lemma \ref{lemma:1} by treating the second problem $x_2$ as a compoition of $k-1$ problems $x_2...x_k$. This changes nothing in the proof. Then, in the proof of theorem \ref{theorem:1}, we simply need to sum over the correct solutions to the compositional problem, for which lemma \ref{lemma:1} holds, and use the same union bound technique.

\section{Probability Bound for Noisy Decoding}\label{probability_bound}
In the presence of logit noise, the probability of each token in a decoding step is changed to:
\begin{equation}
    P(i|context)\rightarrow P'(i|context)\leq \frac{P(i|context)}{P(i|context)+(1-P(i|context))e^X}
\end{equation}
Where $X$ is defined as in equation \ref{eq:weighted_noise}.
The proof follows the proof of lemma \ref{lemma:1} from the text before equation \ref{eq:noisy_logits} up to equation \ref{eq:noise_logit_bound}. The main idea is to write the probability as a softmax over the noisy logits, then extract the original logits from the noise, bound the change in logits due to the noise using Jensen's inequality, and obtaining a bound in terms of the probability without noise.

The advantage of using this form of bound, is that it is relatively succinct, as it removes the dependence on the exact projection of noise on each token in the vocabulary, and instead takes into account the average noise on the different tokens. This allows to efficiently bound the change in probabilities of full sequences due to the noise.

\section{Experimental Details}\label{experimental_details}

\paragraph{Composite Problem Construction:} The composite problems were created by pairs of problems in the formats described in subsection \ref{generation_complexity_experiment}. Our results were based on 50 such composite problems for each experiment. The non-composite problems from human eval and code contests were also tested independently, and we used problems with standalone pass rates $\geq 0.1$, in order to avoid sampling too many solutions in the composite problems (which would typically have accuracy smaller than the product of pass rates of the standalone problems).

\paragraph{Code Generation:} For each problem, code was generated by sampling at $T=1$, and nucleus sampling with $p=0.95$.

\paragraph{Evaluation of Generated Code:} To evaluate the correctness of code generated by the LLM in the experiment described in subsection \ref{generation_complexity_experiment}, we tested the code on the test cases provided in the datasets.

\paragraph{Format of synthetic solutions to composite problems:} In subsections \ref{exponential_length_experiment} (exponential length dependence) and \ref{exp:assumptions} (assumptions), we performed a forward pass of the problem+solution in a scenario with and without composition in order to compare the logits in the two cases. To create correct solutions to the problems, that are ``neutral" (not more likely to be generated by the LLM in a compositional problem than in the standalone case, or vice versa), we created solutions to the composite problems from the standalone problem solutions provided in the datasets. Typically, the model attempted to solve the problems sequentially, either by building functions to solve each problem in the pair, and apply the function sequentially, or by writing the explicit solutions one after the other. We used both templates to create solutions for these experiments.

\paragraph{Sequence Probability Calculation:} In subsection \ref{exponential_length_experiment}, we measured the probability of solutions to problems with composition vs without composition. As explained above, we used to templates for the calculation that are similar to the model's generations (solution in a functional form, where a function is defined for each problem, and in a non-functional form, where the solutions to the problems are written sequentially). In both templates we observed similar trends. In order to avoid an artificial difference between composition and non-composition in the sequence probabilities due to the templates, we measured the probability of sequences after the first few tokens, so that the model has ``time" to adjust to the format in both cases (composition and non-composition), and only measure the probabilities of the actual solution. We did this for both solutions of the compositional problem, for a more fair comparison between composition and non-composition sequence probabilities.

\paragraph{Logit Noise Experiment:} In subsection \ref{exp:assumptions}, we used the templates as mentioned in the above on the sequence probability calculation. We extracted the of the logits of the solutions both in the case of composition and non-composition, and calculated the logit noise as defined in subsection \ref{exp:assumptions} and equation \ref{eq:weighted_noise}.

\paragraph{Additional Human Eval Template:}
In addition to the template presented in the main text, we also tested the following template for Human Eval -- Problem 1 has a True/False output, Problem 2 has an arbitrary output. Composition is to solve problem 1, then if its output is true, print the second problems' output, otherwise, print ``-1":

    \begin{figure}[h!]
        \centering
        \includegraphics[width=0.7\linewidth]{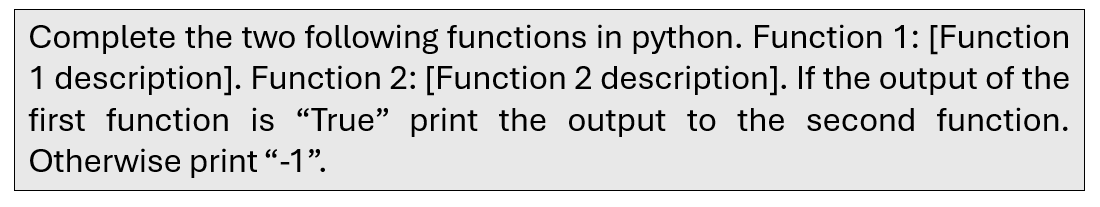}
        \label{fig:template_2}
    \end{figure}

\section{Experiment on Mistral-7B-Instruct-v0.3}\label{sec:mistral}
In order to show the results apply for different models, we perform the same experiment as in subsection \ref{generation_complexity_experiment}. As can be seen, a similar effect is observed, where the number of generations required in composition is typically 10 times more than in the case of generating solutions to both problems in separate contexts.
\begin{figure}[h!]
    \centering
    \includegraphics[width=0.7\linewidth]{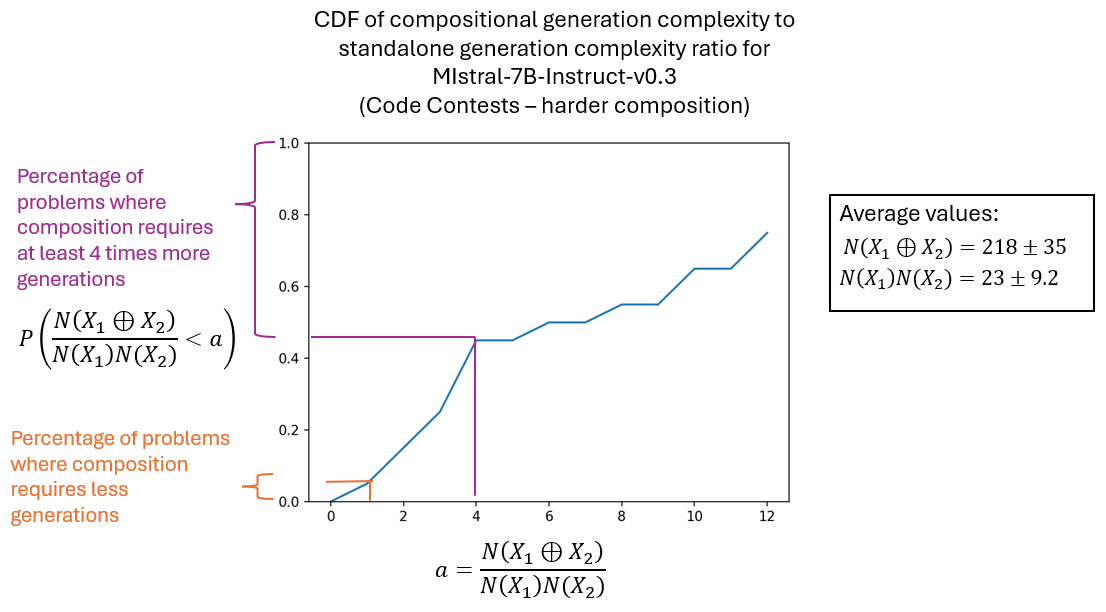}
    \caption{Cumulative distribution function for the ratio of generation complexity using composition, $N(P,x)$, to product of generation complexities for the standalone problems, $N(P,x_1)\cdot N(P,x_2)$ (corresponding to the multi-agent generation complexity). The x axis denotes values for the ratio of generation numbers required to solve the problem in the two cases (composition vs multi-agent), the y axis is the percentage of problems in which the ratio is no larger than this value (\eg~for $a=5$, the y axis value is the percentage of problems where composition requires up to $\times 5$ more samples than the multi-agent case). As can be seen in most of the cases, composition requires twice more samples, and for some problems 10 times more samples.}
    \label{fig:mistral}
\end{figure}

\clearpage
\section{Experiment on Llama-3-70B-Instruct}\label{sec:70B}

\begin{figure}[h!]
    \centering
    \includegraphics[width=0.7\linewidth]{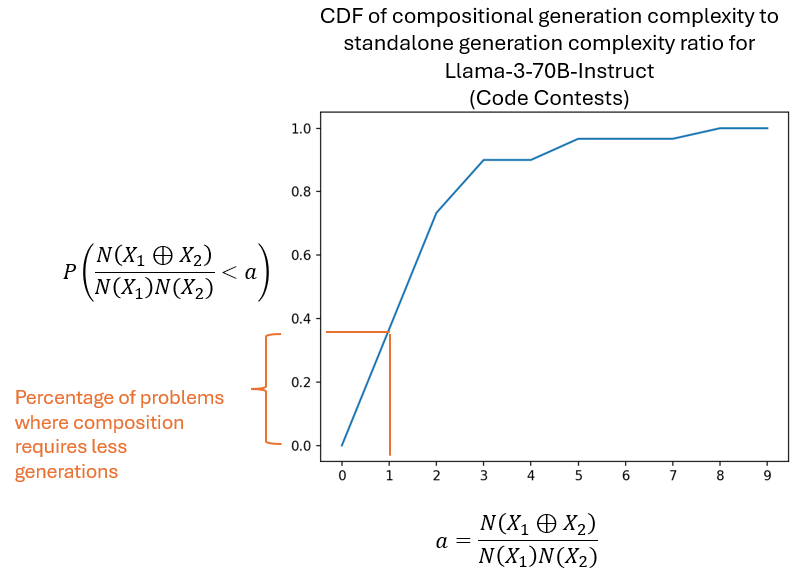}
    \caption{Cumulative distribution function for the ratio of generation complexity using composition, $N(P,x)$, to product of generation complexities for the standalone problems, $N(P,x_1)\cdot N(P,x_2)$ (corresponding to the multi-agent generation complexity). The x axis denotes values for the ratio of generation numbers required to solve the problem in the two cases (composition vs multi-agent), the y axis is the percentage of problems in which the ratio is no larger than this value (\eg~for $a=5$, the y axis value is the percentage of problems where composition requires up to $\times 5$ more samples than the multi-agent case). As can be seen in most of the cases, composition requires twice more samples, and for some problems 10 times more samples.}
    \label{fig:cdf_70B}
\end{figure}

In order to test the dependence of compositional hardness on model size, we repeat the experiment comparing generation complexity with vs without composition (described in subsection \ref{generation_complexity_experiment}).
As seen in figure \ref{fig:cdf_70B}, in most cases, composition requires more generations relative to the non-compositional case, but the ratio is typically smaller than before.

However, if we increase the difficulty by concatenating four problems in the context and ask the model to only solve two of them (which keeps the compositional problem effectively a concatenation of two problems):
        \begin{figure}[h!]
        \centering
        \includegraphics[width=0.7\linewidth]{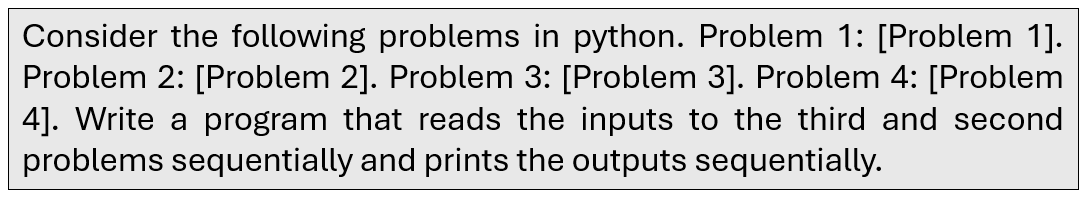}
        \label{fig:template_4}
    \end{figure}

\begin{figure}[h!]
    \centering
    \includegraphics[width=0.9\linewidth]{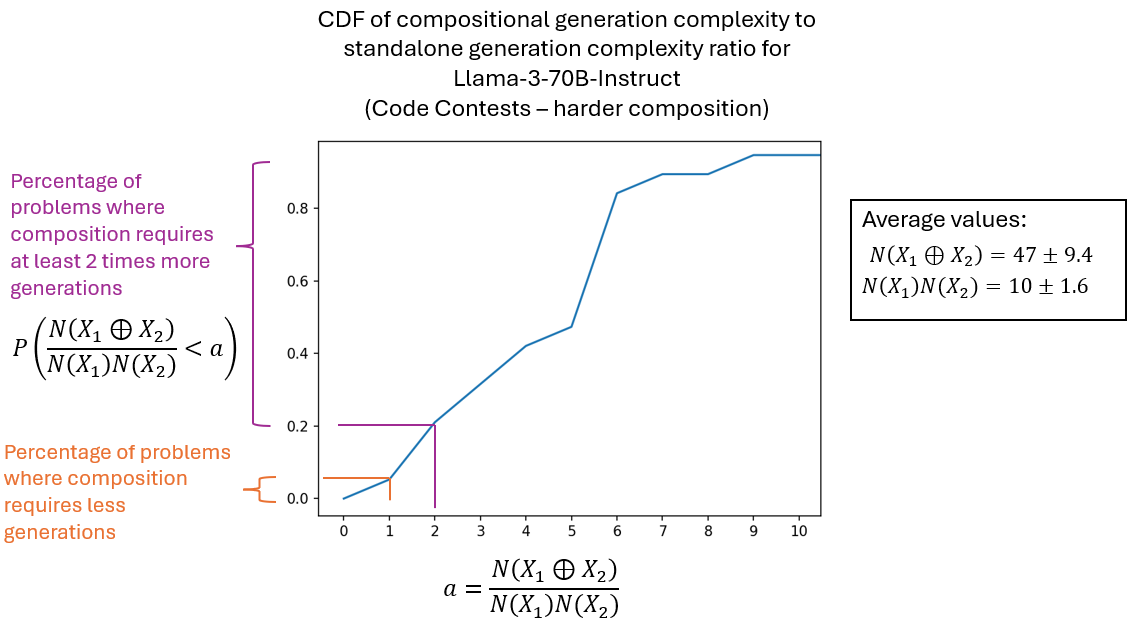}
    \caption{Cumulative distribution function for the ratio of generation complexity using composition, $N(P,x)$, to product of generation complexities for the standalone problems, $N(P,x_1)\cdot N(P,x_2)$ (corresponding to the multi-agent generation complexity). Here we explicitly ask the model to concatenate the solution to two problems, but also expose it to two additional ones that it is not meant to solve. The x axis denotes values for the ratio of generation numbers required to solve the problem in the two cases (composition vs multi-agent), the y axis is the percentage of problems in which the ratio is no larger than this value. As can be seen in most of the cases, composition requires at least 5 times more samples.}
    \label{fig:cdf_70B_harder}
\end{figure}

We once again obtain large ratios of compositional generation complexity, as seen in figure \ref{fig:cdf_70B_harder}. This demonstrates the model's difficulty in extracting only the relevant information within the context for solving the problems (even when the separation is explicit), and that the noisy context increases the composition difficulty, such that even a concatenation of two problems becomes significantly more difficult. This is in accordance with our theory, where we consider a problem $x$, which is implicitly decomposable into two problems $x_1,x_2$, whose concatenated solutions, $y_1\oplus y_2$, solves the problem. We obtain that the generation complexity to $x$, is much higher than the product of generation complexities for $x_1$ and $x_2$ due to the noise from the context.

\clearpage

\section{Numerical Estimation of $\Delta$ and $\sigma$}\label{sec:delta_sigma_estimation}

\begin{figure}[h!]
    \centering
    \includegraphics[width=0.75\linewidth]{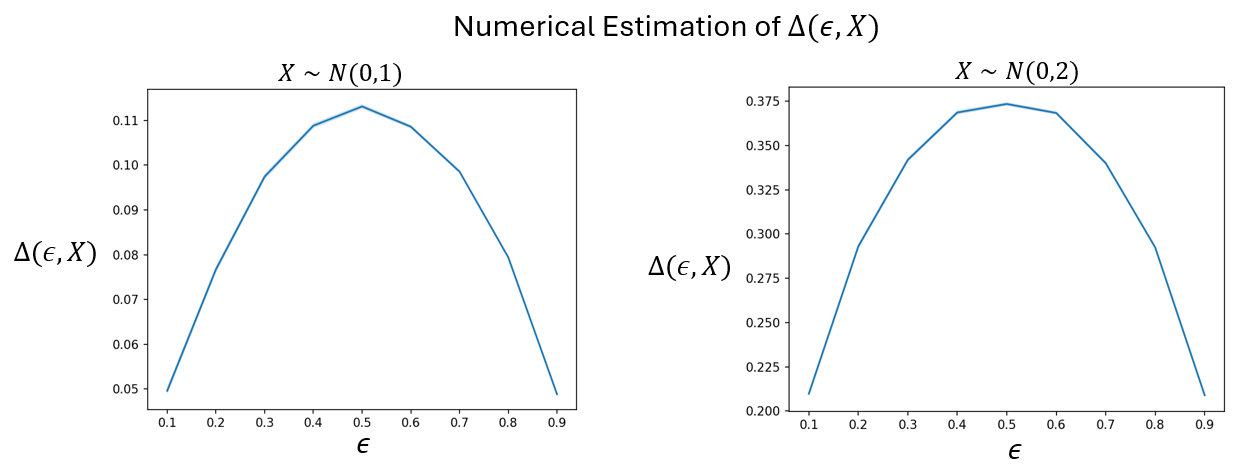}
    
  \includegraphics[width=0.75\linewidth]{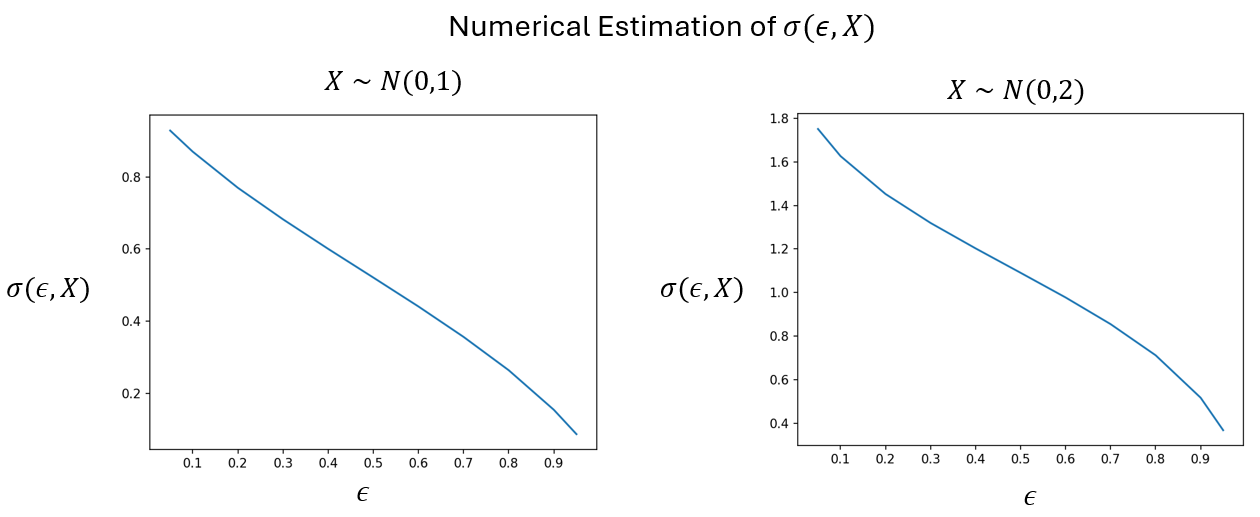}
  
    \caption{Numerical Estimation of $\Delta(\epsilon,X)$ and $\sigma(\epsilon,X)$.}
    \label{fig:delta_numerical}
\end{figure}

To estimate $\Delta(\epsilon,X)$ we approximate the noise as a Gaussian with mean 0 and try two values standard deviation $\sigma=1$ and $\sigma=2$, then calculate $\Delta(\epsilon,X)\approx mean_{\{X_i\}}[\epsilon+(1-\epsilon)e^{X}]$. The values are presented in figure \ref{fig:delta_numerical}, and match the estimation of $\Delta$ in the range of $0.05$ to $0.2$ from the above subsection. Similarly, we estimate $\sigma(\epsilon,X)$. The values are plotted in figure \ref{fig:delta_numerical}, showing typical values of $\sigma\approx 1- 2$

\end{document}